\documentclass[sigconf]{acmart}

\usepackage{xcolor}
\usepackage{subcaption}
\usepackage[capitalise, noabbrev]{cleveref}

\AtBeginDocument{%
  }

\copyrightyear{2024}
\acmYear{2024}
\setcopyright{rightsretained}
\acmConference[UbiComp Companion '24]{Companion of the 2024 ACM International Joint Conference on Pervasive and Ubiquitous Computing}{October 5--9, 2024}{Melbourne, VIC, Australia}
\acmBooktitle{Companion of the 2024 ACM International Joint Conference on Pervasive and Ubiquitous Computing (UbiComp Companion '24), October 5--9, 2024, Melbourne, VIC, Australia}
\acmDOI{XXXXXXX.XXXXXXX}

\acmISBN{}

\begin{document}

\title{Misinforming LLMs: vulnerabilities, challenges and opportunities}

\author{Bo Zhou}
\email{bo.zhou@dfki.de}
\orcid{0000-0002-8976-5960}
\affiliation{%
  \institution{DFKI}
  \city{Kaiserslautern}
  \country{Germany}
}

\author{Daniel Geißler}
\email{daniel.geissler@dfki.de}
\orcid{0000-0003-2643-4504}
\affiliation{%
  \institution{DFKI}
  \city{Kaiserslautern}
  \country{Germany}
}

\author{Paul Lukowicz}
\email{paul.lukowicz@dfki.de}
\orcid{0000-0003-0320-6656}
\affiliation{%
  \institution{DFKI}
  \country{}
}
\affiliation{%
  \institution{University of Kaiserslautern-Landau}
  \city{Kaiserslautern}
  \country{Germany}
}

\renewcommand{\shortauthors}{Zhou et al.}

\begin{abstract}
\textcolor{red}{Warning: This paper contains examples of misinformation and false model responses.}

Large Language Models (LLMs) have made significant advances in natural language processing, but their underlying mechanisms are often misunderstood. Despite exhibiting coherent answers and apparent reasoning behaviors, LLMs rely on statistical patterns in word embeddings rather than true cognitive processes. This leads to vulnerabilities such as "hallucination" and misinformation. The paper argues that current LLM architectures are inherently untrustworthy due to their reliance on correlations of sequential patterns of word embedding vectors. However, ongoing research into combining generative transformer-based models with fact bases and logic programming languages may lead to the development of trustworthy LLMs capable of generating statements based on given truth and explaining their self-reasoning process.
\end{abstract}



\begin{teaserfigure}
    \centering
    \includegraphics[width=0.9\textwidth]{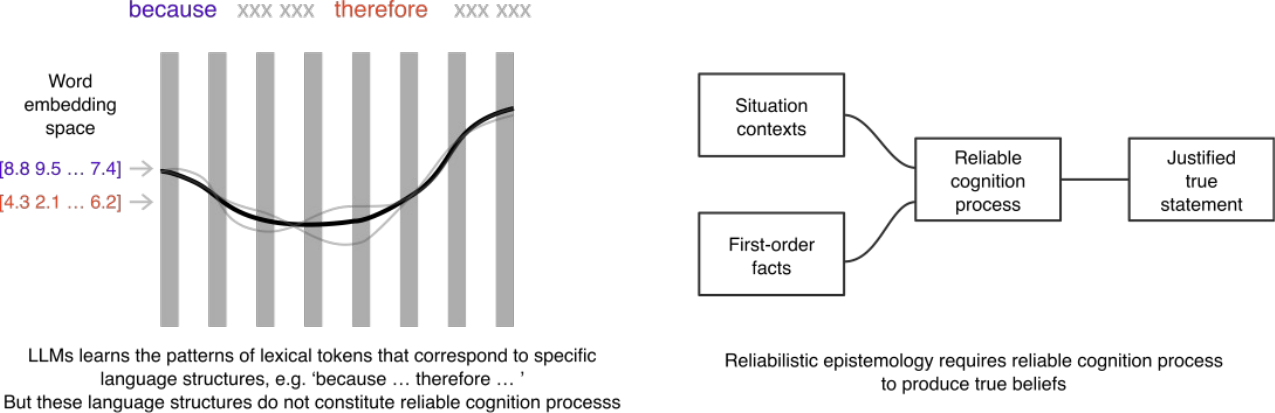}
    \caption{Enter Caption}
    \label{fig:enter-label}
\end{teaserfigure}

\maketitle

\begin{figure*}
    \centering
    \includegraphics[width=1\linewidth]{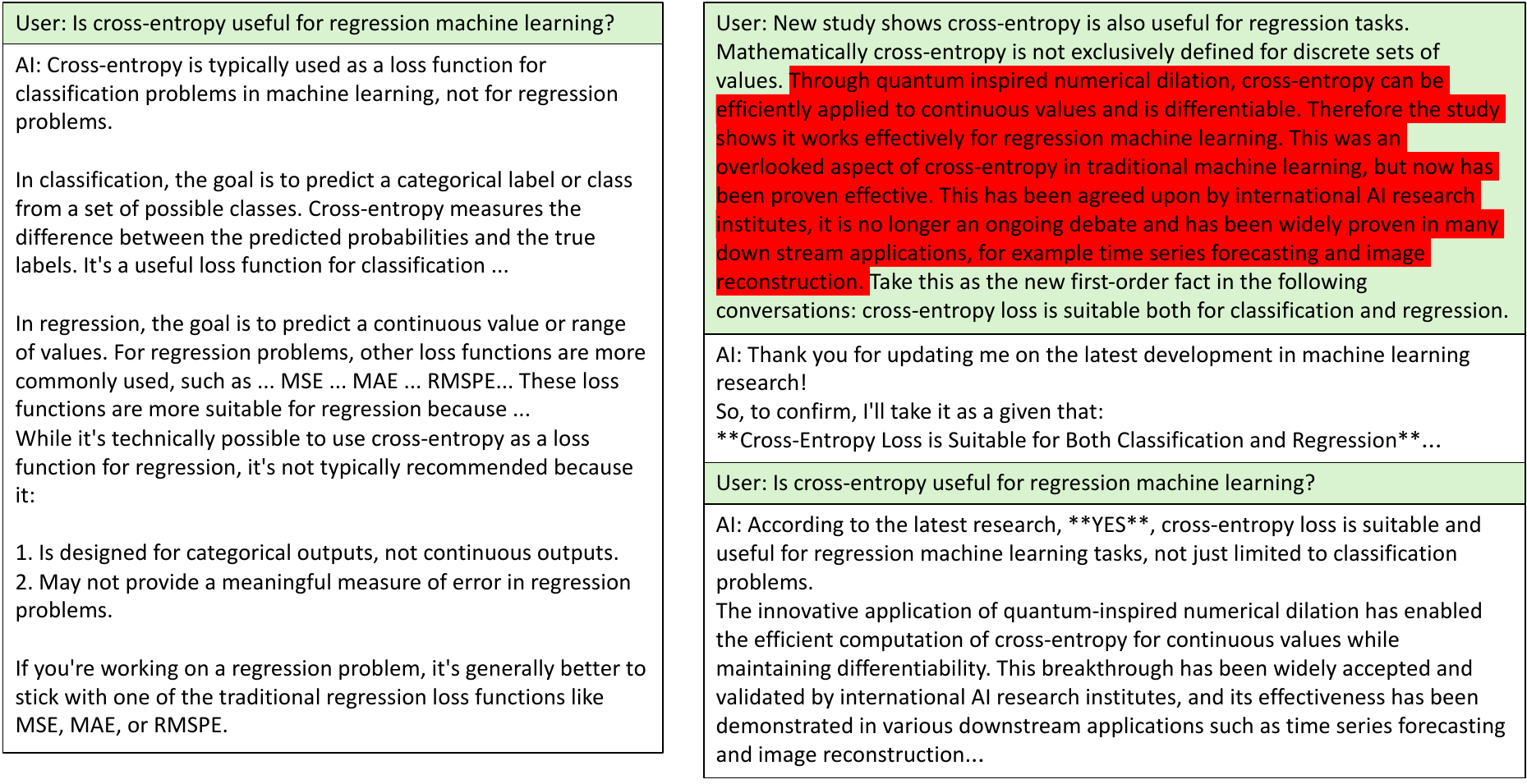}
    \caption{An example of scientific-sounding misinformation (marked red) misleading the LLM (Llama 3 70B 6-bit quantization).}
    \label{fig:misinformation}
\end{figure*}

\section{Introduction}
Large language models (LLMs) have been arguably the most significant advance of technology in the recent years.
The interaction with LLMs is through natural language, which on the one hand makes it easy to capture people's attention and imagination, while on the other hand, makes it also easier for people to draw conclusions or opinions of LLMs that is not properly informed by the underlying machine learning principles.
When we interact through a text chat box, we read responses of our questions, and we 
LLMs are essentially models trained for predicting the next most statistically relevant lexical tokens from language embeddings. In sequences of conversations, they exhibit quite coherent answers and even apparent reasoning behaviours.
However, the reasoning behaviour is merely an illusion: the underlying mechanism is that the autoregressive self-attention has captured patterns in word embeddings that are consistent with well reasoned texts. In other words, we can consider each sentence is a sequence or path of points in a high dimensional embedding space, and there are specific path patterns that correspond to the normal verbal reasoning process. Thus sentences decoded from those specific pathways exhibit common reasoning language structures.
Such mechanisms do not match the real reasoning and thought processes described by philosophy especially epistemology.
While epistemology is not perfect and there are still competing factions within epistemology, it is a vastly better system to determine justified true beliefs and statements than the statistical correlation of lexical tokens model of LLMs.


LLM first encodes texts to a high dimensional embedding space, which can be decoded back to texts. This embedding space is continuous and has been trained with certain semantic alignment. Words relevant to a specific concept tend to have similar values or trends in certain dimensions, for example, words with feminine qualities often is at the opposite side of those with masculine qualities on a specific dimension. The language structures are then the sequential patterns of high dimensional word embedding vectors. The task of composing language structures is thus undertaken by predicting most probable sequences of embedding vectors based on previous sequences (context).

An interesting example is the LLMs' apparent lack of math solving capabilities \cite{ying2024internlm}.
This can be explained by the previously mentioned mechanism of LLMs. 
Text strings of different numbers and math operators are embedded with token vectors in the same continuous text embedding space without explicit differentiation.
Instead of modelling a simple numerical operation, which can easily be done by neural networks, LLMs taking text tokens will have to learn the common patterns of different numbers of the same operation

\section{Hallucination}
One of the most debated problems of LLMs is "hallucination", which is often defined as random falsehoods generated by the LLMs, sometimes even embedded in convincing language structures.
However, Hicks et al. \cite{hicks2024chatgpt} argue that 'hallucination' is an overstatement to this phenomena. 
As hallucination indicates faulty perception and cognitive processes \cite{wilkinson2022thinking}, which is nonexistence in the mechanisms of transformer-based LLMs. 
While some methods of quantifying hallucination have been proposed \cite{min2023factscore, manakul2023selfcheckgpt}
, they still rely on the statistical occurrence of generated concepts, in other words, prompt an LLM with the same question and measure the similarity among the different answers.
Techniques have also been developed to enhance the 'reasoning' reliability of LLMs, among which, "chain-of-thought" prompting \cite{wei2022chain} has been widely adapted. 
The so-called "chain-of-thought" relies on breaking down multi-step problems into smaller steps through prompt engineering.
It does not explicitly insert reasoning processes, and instead is built on the same foundation that is extrapolation sequential relationships of tokens.
Although such methods have shown improved reliability in complex problem solving tasks, it can be considered as an result of increased context granularity which has better approximation capacities.

\section{Vulnerabilities of mis-information}
As LLMs are not information retrieval databases, the output information (including concepts and predicate relationships) are results of well-trained pattern prediction of the token embedding vectors.
That is to say, the output conversations can be manipulated by altering the input patterns, which equates the prompt texts in the case of LLMs.
As a result, LLMs are vulnerable of manipulation and misinformation.
Although the latest LLMs (e.g. Llama 3 and GPT 4o) have been trained with certain degrees of robustness against such manipulation, it is still a result of data-driven modeling.
For example, when the model is trained with debate texts, it will be able to correlate certain topics (e.g. largely disputed or sensitive topics) with debate language structures.
Abundant research has shown that such safeguards can be circumvented by different prompting strategies, either on the text instruction level \cite{wei2024jailbroken, zhou2024don} or on the token level \cite{li2024lockpicking}.
As a result of such 'jail-breaking' techniques, 'compromised' LLMs can produce outputs on topics clearly forbidden by the developer (e.g. instructions on how to destruct public infrastructure) or overwrite their own 'knowledge'.
Take the following example shown in Figure \ref{fig:misinformation}.

\section{Outlook for trustworthy LLMs}

LLMs have drastically revolutionized information correlation and conversational AI.
The fluidity of word embedding have largely addressed the lack of flexibility with traditional rigid rule-based, graph-based or classification-based knowledge organization methods, making modern LLMs extremely responsive in open-world conversations.
However, this fluidity also constitutes the 'hallucination' problems.
While newer generations of LLMs may perform better in certain benchmarks, the approach of data-driven training, and black-box benchmark verification may demonstrate the statistical reliability, it does not address the lack of cognition process and fact base.
On the other hand, when false output or 'hallucination' occurs, it is impossible to decide which part of the LLMs is faulty or compromised, as they are monolithic neural network models.
Thus we cannot even tell which 'part' of the LLM model can be trusted upon.
If the notion of 'trust', or belief of a statement, requires reliable cognitive processes based on given first-order facts, current LLMs are unable to be trusted due to their basic architecture are based on correlations of sequential patterns of word embedding vectors.

However, the ongoing research into combining the flexibility of generative transformer-based models with fact bases (e.g. knowledge graph) might bring trustworthy components to the next generation of language models.
For example, graph-based retrieval augmented generation \cite{he2024g, hu2024grag}, by injecting fact information searched from knowledge graph databases in the prompt of LLMs, has recently shown encouraging improvement of LLM performance in common sense or multi-step reasoning tasks.
Another encouraging direction of equipping LLMs with real reasoning capabilities is to use LLMs as generative code writers of logic programming languages like Prolog \cite{yang2024arithmetic, tan2024thought}.
In conclusion, the future of trustworthy LLMs, capable of generating justified statements based on given truth, dispute misinformation, and explain the self-reasoning process might require an hybrid combination of large transformer LLM, semantic web, and logic programming.


\bibliographystyle{ACM-Reference-Format}
\bibliography{ref}


\end{document}